\documentclass[conference]{IEEEtran}
\IEEEoverridecommandlockouts
\usepackage{cite}
\usepackage{amsmath,amssymb,amsfonts}
\usepackage{algorithmic}
\usepackage{graphicx}
\usepackage{caption}
\usepackage{subcaption}
\usepackage[linesnumbered,ruled]{algorithm2e}
\usepackage{textcomp}
\usepackage{xcolor}
\usepackage{romannum}
\usepackage{graphicx}
\usepackage{flushend}
\usepackage{multirow}
\usepackage{cleveref}
\def\BibTeX{{\rm B\kern-.05em{\sc i\kern-.025em b}\kern-.08em
    T\kern-.1667em\lower.7ex\hbox{E}\kern-.125emX}}
    
\makeatletter
\newcommand{\linebreakand}{%
  \end{@IEEEauthorhalign}
  \hfill\mbox{}\par
  \mbox{}\hfill\begin{@IEEEauthorhalign}
}
\makeatother

\begin{document}


\title{Variational Autoencoder Learns Better Feature Representations for EEG-based Obesity Classification}
\author{
\IEEEauthorblockN{Yuan Yue}
\IEEEauthorblockA{\textit{Department of Information Science} \\
\textit{University of Otago)}\\
Dunedin, New Zealand \\
yueyu445@student.otago.ac.nz}
\and
\IEEEauthorblockN{ Jeremiah D. Deng}
\IEEEauthorblockA{\textit{Department of Information Science} \\
\textit{University of Otago}\\
Dunedin, New Zealand \\
jeremiah.deng@otago.ac.nz}
\and
\IEEEauthorblockN{Dirk De Ridder}
\IEEEauthorblockA{\textit{Department of Surgical Science} \\
\textit{University of Otago}\\
Dunedin, New Zealand \\
dirk.deridder@otago.ac.nz}
  \linebreakand 
\IEEEauthorblockN{Patrick Manning}
\IEEEauthorblockA{\textit{Department of Medicine} \\
\textit{University of Otago}\\
Dunedin, New Zealand \\
patrick.manning@otago.ac.nz}
\and
\IEEEauthorblockN{Divya Adhia}
\IEEEauthorblockA{\textit{Department of Surgical Science} \\
\textit{University of Otago}\\
Dunedin, New Zealand \\
divya.adhia@otago.ac.nz}
}

\maketitle

\begin{abstract}
Obesity is a common issue in modern societies today that can lead to various diseases and significantly reduced quality of life. Currently, research has been conducted to investigate resting state EEG (electroencephalogram) signals with an aim to identify possible neurological characteristics associated with obesity. In this study, we propose a deep learning-based framework to extract the resting state EEG features for obese and lean subject classification. Specifically, a novel variational autoencoder framework is employed to extract subject-invariant features from the raw EEG signals, which are then classified by a 1-D convolutional neural network. Comparing with conventional machine learning and deep learning methods, we demonstrate the superiority of using VAE for feature extraction, as reflected by the significantly improved classification accuracies, better visualizations and reduced impurity measures in the feature representations. 
Future work can be directed to gaining an in-depth understanding regarding the spatial patterns that have been learned by the proposed model from a neurological view, as well as improving the interpretability of the proposed model by allowing it to uncover any temporal-related information. 
\end{abstract}


\section{Introduction}
Obesity is a worldwide health problem today that is associated to the dysfunction of various body systems including the heart, the liver, kidneys, joints, and the reproductive system~\cite{buechler_adiponectin_2011,silva_obesity_2017,verma_obesity_2017,wolin_obesity_2010}. It is also the cause of various diseases such as type 2 diabetes, cardiovascular diseases and cancers~\cite{wlodarczyk_obesity_2019}.

While many researchers are investigating obesity-related clinical characteristics, attention has been increasingly paid to the effect of obesity on the neurological perspective~\cite{sui_obesity_2020,lowe_prefrontal_2019}. Studies have shown that obesity is associated with cognitive impairment and altered structure of various brain networks~\cite{difeliceantonio_dopamine_2019}. For example, it is suggested that the abnormal brain connection in the somatosensory cortex and insula areas of obese individuals make them less capable of predicting the energy need, thus, tending to overtake food~\cite{yue_finding_2022}. On the other hand, it is found that an altered hippocampal structure, which highly correlates to the Alzheimer’s type dementia,  is observed in obese individuals~\cite{obrien_neurological_2017}.

As a non-invasive technique for brain activity recording, EEG technology has been extensively used to study various brain activity patterns, including obesity-related neurological patterns~\cite{bethge_eeg2vec_2022,babiloni_classification_2016,allison_braincomputer_2007}. For instance, in~\cite{imperatori_modification_2015}, EEG data are collected from a group of healthy participants and obese participants. Traditional statistical analysis was then applied to investigate the difference in brain activities between obese and healthy people.  Their results suggest that psychopathological mechanisms that are similar to substance-related disorders and addictive disorders are observed in obese brains. These mechanisms, which include an increased spectral power or functional connectivity within a certain frequency range, suggest a different activation mechanism of the cognitive and emotional process when exposing to an environment which has food-related cues between healthy people and obese people.

On the other hand, substantial progress has been made in the machine learning and deep learning field in recent decades. This has significantly improved the efficiency of analyzing any EEG-related tasks. However, to our best knowledge, few studies have applied machine learning to study obesity-related brain activities. A traditional machine learning approach was proposed in~\cite{yue_finding_2022} to identify distinctive EEG patterns of obese individuals, where 
source localisation was performed to select the region of interest, and  functional connectivity features were used to train a classifier based on a support vector machine (SVM) with radial basis function (RBF) kernels. An average classification accuracy of 0.912 was reported. 

\begin{figure*}[h] 
    \centering
    \includegraphics[width=0.99\textwidth]{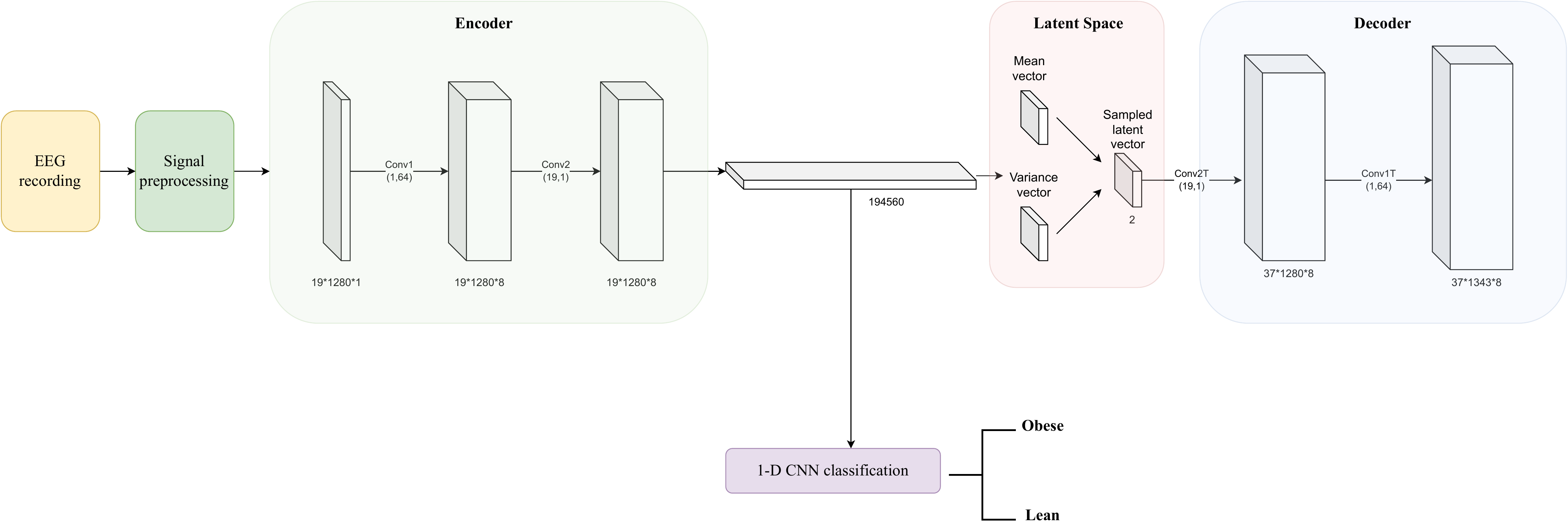}
    \caption{{Overall process of the proposed work.}}
    \label{fig:flowchart}
\end{figure*}

In this paper, we explore obese brain activities by examining raw EEG data recorded in resting state using a deep learning approach. The novelty of our study includes:
\begin{itemize}
  \item Our study is the first one that focuses on identifying obese brain activities by examining resting state EEG data using a deep learning-based framework;
  \item  We demonstrated the superior effectiveness of using a Variational Autoencoder (VAE) model to extract features in resting state EEG classification tasks; 
  \item In additional to favourable visualization outcomes, we proposed a quantitative measure based on impurity to evaluate the separability of the feature space, where the superiority of VAE features was further confirmed. These added to the explainability of our proposed model. 
    \end{itemize}
The overall process of our study is demonstrated in Fig ~\ref{fig:flowchart}. After data collection and preprocessing, we first use an unsupervised VAE to learn meaningful feature representations (i.e. the output of the encoder), a 1-D convolutional neural network (CNN) is then applied to perform the classification task using the feature representation as input. 

The rest of the paper is structured as follows:
In Section~\ref{sec:related_work} we briefly introduce the related works and terminologies that will be referred to in our study. The proposed model and detailed experiment procedures including data description, the implementation of the proposed model, as well as the classification process, are discussed in Section~\ref{sec:method}. Then, in Section~\ref{sec:result} we evaluate the performance of the proposed model, and compare it with baseline models. Furthermore, we present the outcome of visualization and impurity measures on separability of the acquired latent features. The paper is concluded in Section~\ref{sec:conc}.  

\section{Related Work}
\label{sec:related_work}
\subsection{Variational Autoencoder (VAE)}

Autoencoders are neural networks that can learn features from the input data in an unsupervised way~\cite{pathirage_structural_2018}. It consists of two main parts: an encoder and a decoder. Both parts have a similar structure to a CNN without the final classification layer. The encoder
compresses the input data and learns useful features through non-linear dimension reduction in the latent space, whereas the decoder decompresses the features to reconstruct the input data. The objective function of an autoencoder can be 
 formally written as: 
\begin{equation}
 \mathcal{L}_{\mathrm{AE}}=E(\|\mathbf{x}-\hat{\mathbf{x}}\|^2),
\end{equation} where $\mathbf{x}$ is the input data sample and $\hat{\mathbf{x}}$ is the corresponding reconstruction of the input data sample.

The biggest drawback of autoencoder is its strong tendency to overfit~\cite{steck_autoencoders_2020}, as it is solely trained to encode and decode with as little  loss as possible regardless of how the latent space is organized. To address this problem and to turn an autoencoder into a generative model, VAE has been developed~\cite{Kingma2014} and found as effective solutions~\cite{steck_autoencoders_2020,bi_eeg-based_2019}. The aim of VAE is to ensure that the latent space is regular enough, therefore the training process can be regularized to avoid overfitting. 
To achieve this regularization in the latent space, VAE modified the traditional autoencoder by encoding an input data sample as a distribution in the latent space. The decoder then performs the decompressing process using data points that have been sampled from the latent space. Moreover, the regularization in the latent space is implemented by enforcing the encoded latent variables toward a Gaussian distribution. 
Thus, the objective function of a VAE contains two parts: minimizing the reconstruction error which makes the encoding-decoding process as precise as possible, and forcing the distribution outputted by the encoder as close as possible to a Gaussian distribution. The loss function of a VAE can be formally written as:
\begin{equation}
    \mathcal{L}_{\mathrm{VAE}}=E(\|\mathbf{x}-\hat{\mathbf{x}} \|^{2}) + D_\mathrm{KL}(q(\mathbf{z}|\mathbf{x})|| p(\mathbf{z})),
\end{equation} 
where $\mathbf{z}$ is the latent variable, $q(\mathbf{z}|\mathbf{x})$ is the latent distribution generated by the encoder, $p(\mathbf{z})$ is the prior distribution in the latent space which follows a Gaussian distribution, and $D_\mathrm{KL}(.)$ denotes the Kullback-Leibler divergence.

VAE has been increasingly used in EEG classification tasks to learn robust features \cite{bi_eeg-based_2019,bethge_eeg2vec_2022,bollens_learning_2022}. For instance,~\cite{lee_motor_2022}  proposed a channel-wised VAE-based convolutional neural network to classify motor imagery dataset that has 4 classes. The VAE here comprises 1 input layer, 5 hidden layers that follow a deep-wise convolution structure, and 1 fully connected layer. By employing the channel-wised VAE model, an average classification accuracy of 0.83 is achieved comparing to the accuracy score of 0.69 achieved by using the basic EEGNet.

\subsection{Convolutional Neural Network (CNN)}
A CNN usually consists of multiple convolution-pooling layer pairs followed by a fully connected layer. Layers with other functions such as the batch normalization layer and activation layer can be added depending on the modeling needs. 
The convolution layer is where the majority of computation has been performed. Here, one or more kernels (i.e., filters) slide over the input data passing by the previous layer. For each filter, the dot product (i.e., feature map) between the input data and the kernel is then computed.
Next, the learned feature maps are passed to a pooling layer. In this layer, the number of parameters is reduced by taking the average or the maximum of each of the $N$ parameters ($N$ is a user-defined integer greater than 0). 
Finally, the pooled features are fed into a fully connected layer, at which each of the input is connected to the activation unit and the final class of each data sample can be predicted. 

\subsection{EEGNet}
\begin{figure*}[!t] 
    \centering
    \includegraphics[width=0.9\textwidth]{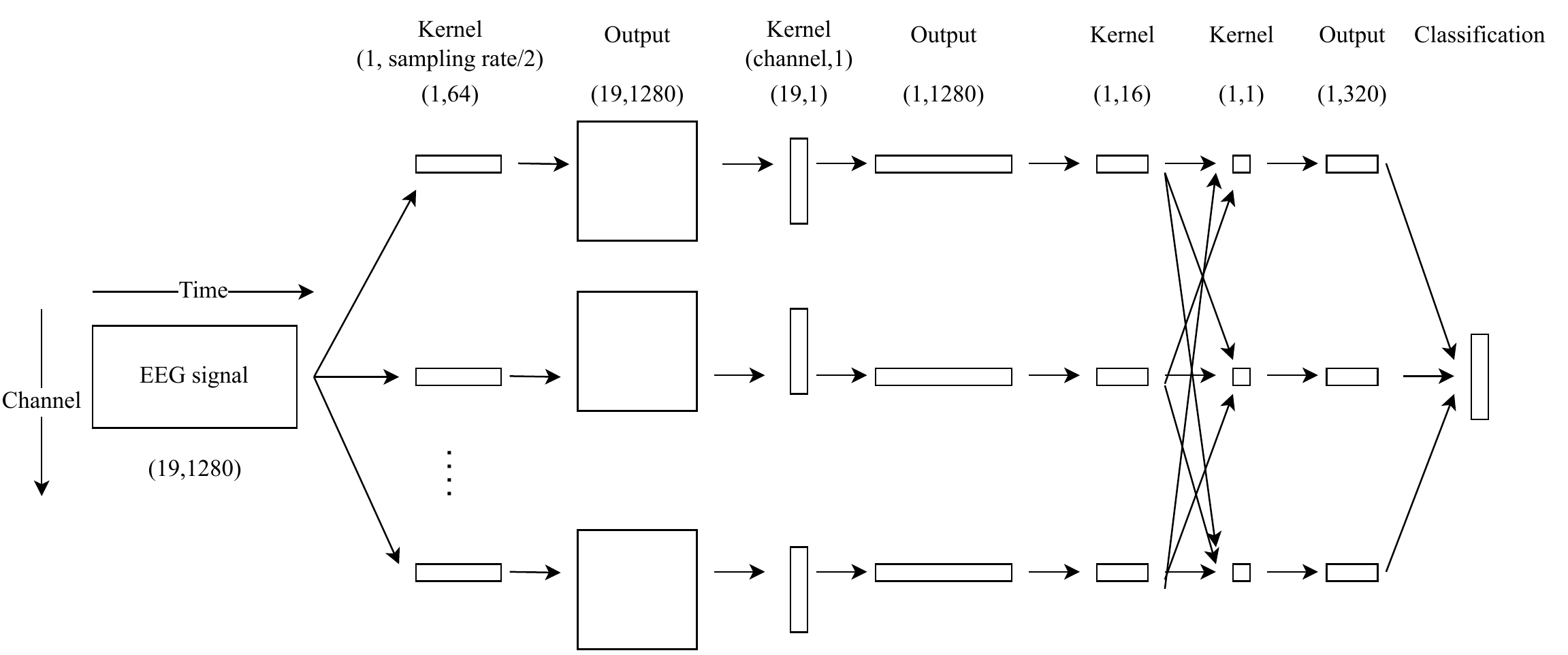}
    \caption{{Architecture of EEGNet.}}
    \label{fig:eegnet_arch}
\end{figure*}
EEGNet~\cite{lawhern_eegnet_2018} is an extensively used deep learning model for EEG classification tasks due to its advance in learning various EEG representations~\cite{deng_advanced_2021,bethge_eeg2vec_2022,zhu_eegnet_2021}, as well as its capability of achieving a relatively good performance while requiring low computation cost. Many studies have performed EEG classification tasks by employing EEGNet-based models with task-specific customization added~\cite{bethge_eeg2vec_2022,raza_single-trial_2020,deng_advanced_2021}. For example, in~\cite{raza_single-trial_2020}, neural structured learning (NSL) is combined with EEGNet to classify binary motor imagery dataset. Their model allows the relational information and structural similarities contained in the input signal to be well maintained in the training process. Regarding their experiment results, the basic EEGNet model demonstrated its efficiency by achieving a classification accuracy of 0.722 when it is applied alone, with a small increment of 0.039 after adding the NSL. 

The architecture of EEGNet, as well as the parameter size adapted in our study, are demonstrated in Fig. ~\ref{fig:eegnet_arch}. To simplify the visualization and to highlight the main function of EEGNet, we showed only the main convolution layers contained in the model, without including batch normalization layers, average pooling layers, and activation layers. EEGNet consists of a temporal filtering layer at which a kernel size equal to $1\times \mathrm{(Sampling\ Rate/2)}$ is applied, followed by a deep-wise spatial filtering layer at which a kernel size of $\mathrm{(\#\ of\ channels)}\times 1$ used. A separable convolution block is then added. The adapted deep-wise separable convolution structure allows EEGNet to achieve a high computation efficiency by significantly reducing the number of parameters. The final class of each sample is then computed by a linear fully-connected layer. 
 
\subsection{Cross-subject variability}
Cross-subject variability refers to the variation in brain activities among different individuals, as each person has a unique brain anatomy and functionality~\cite{gu_what_2014}.  In EEG analysis tasks, subject-dependent information contained in the input signal tends to compromise model performances since the model has been trained on task-unrelated whereas subject-specific information~\cite{lee_motor_2022,hang_cross-subject_2019}. To address this problem, current research aims to find task-related as well as robust brain activity features that are consistent across all individuals~\cite{bollens_learning_2022,lee_motor_2022}. Many approaches have been developed for this purpose. For examples, a semi-supervised model~\cite{bollens_learning_2022} was used to learn subject-independent featuresm abd domain adaptation~\cite{gedeon_reducing_2019} was performed to transfer all subjects’ data into the same latent space for further classification. 

\section{Method}
\label{sec:method}
\subsection{Data description}
The EEG datasets used in this study were collected from 30 obese females and 30 lean (i.e., healthy) females who are between 25 to 65 years old. Subjects who have a body mass index (BMI) higher than 30 are defined as obese individuals and those who have a BMI lower than 25 are defined as lean individuals. The international 10–20 system was adopted for EEG recording.  The lab environment and EEG recording devices were controlled to be in the same condition during the recording process. 
For each subject, resting state EEG data are recorded during eye-closing state prior to any meal consumption. 

The first five seconds of each EEG recording were discarded as they usually contain a high level of noise. The EEG recordings were then resampled to 128Hz, and band-pass filtered between 0.1Hz - 45Hz. Next, each EEG signal is segmented into 10-second consecutive epoch (10$\times$128=1280 timepoints). Each epoch is then used as an independent sample.

\subsection{Feature Extraction}
A novel VAE model is developed for feature extraction. As we mentioned above, VAE works by letting the encoder learn a parametrized Gaussian distribution of the input data in the latent space and letting the decoder reconstruct the input data using the latent representation. We therefore can infer that the majority of the information contained in the raw EEG signal can be captured in the encoder's output. Therefore, instead of using the raw EEG data as the input for the classification task, we used the encoder output as the input features.

The detailed architecture of the proposed VAE is shown in Fig. ~\ref{fig:flowchart}. This architecture is designed based on the concept of EEGNet~\cite{lawhern_eegnet_2018}. In the encoding part of the proposed VAE, we first adapted a temporal convolution layer with a kernel size of 1$\times$128/2=1$\times$64, to extract temporal features. We then performed a spatial convolution by using kernels with a size of 19$\times$1, to extract spatial features. Each convolution layer is followed by a batch normalization layer and a leaky ReLU layer. The decoder is then designed by taking the inverse of the encoder. 

We then trained the VAE on normalized raw EEG data of each subject, and the final encoded outputs were used as the input features for later classification.

\subsection{Train and test split}
The dataset was partitioned based on subject-based cross-validation. Within each fold, 6 subjects (3 lean and 3 obese) were held out for testing and 54 (27 lean and 27 obese) subjects were used for training. Within each training fold, 6 subjects (3 obese and 3 lean) were further selected for validation. The final testing score are obtained by averaging the scores of all folds.

\subsection{Classification}
\begin{figure}[!h] 
    \centering
    \includegraphics[width=0.4\columnwidth]{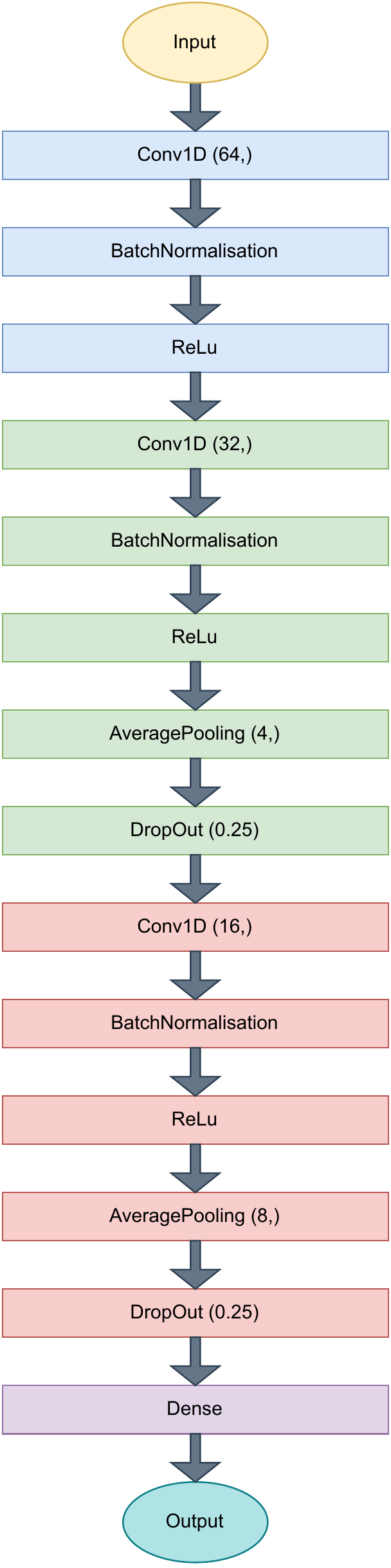}
    \caption{Architecture of the proposed 1-D CNN used for final classification.}
    \label{fig:1dcnn}
\end{figure}

We used a 1-D CNN, a SVM with a RBF kernel, and a multilayer perceptron (MLP) to  predict the label of each data sample, respectively. The architecture of the proposed 1-D CNN is shown in Fig. ~\ref{fig:1dcnn}. In the first and second convolution layers we used 8 filters with a size of 64, and 16 filters with a size of 32, respectively. We then added an average pooling layer with a size of 4, and a drop-out layer with a drop rate of 0.25 to reduce the number of parameters and prevent the model from overfitting. The final convolution layer consists of 32 filters with a size of 16, followed by an average pooling layer with a size of 8 and a drop-out layer with a drop rate of 0.25. After each convolution, batch normalization was performed, and an activation layer was added. In this way, the network can learn and converge at a faster speed through regularization.
Next, since each subject has 26 samples (epochs), we then determined the final label of each subject by counting how many samples that belong to the subject are classified as obese and how many are classified as lean.
The performance of the three classifiers is evaluated in the next section. 

\section{Results and Discussion}
\label{sec:result}
\subsection{Classification performance comparison}
The test scores (i.e., accuracies) obtained by using the 1-D CNN, the SVM and the MLP are listed in Table ~\ref{tab:classfier_result}. The highest classification scores (0.951 and 0.937 for subject level classification and epoch level classification respectively) are obtained when using the 1-D CNN. Therefore, we will only consider its use in the following work on model comparison that involves both the EEGNet and the proposed model using VAE. 

\begin{table}
 \caption{Test scores (mean accuracy $\pm$ standard deviation) obtained using 1-D CNN, SVM, MLP as classifiers.}
 \begin{center}
 \begin{tabular}{ | c |c|c | }
   \hline
   Methods & Subject-level scores & Epoch-level scores  \\ 
   \hline\hline
   1-D CNN  & 0.951 $\pm$ {0.104} & 0.937 $\pm$ {0.130} \\ 
   \hline
   SVM  & 0.917 $\pm$ {0.170} & 0.920 $\pm$ {0.139} \\
   \hline
   MLP & 0.883 $\pm$ {0.299} & 0.898 $\pm$ {0.236} \\ 
   \hline
 \end{tabular}
 \end{center}
 \label{tab:classfier_result}
 \end{table}

\begin{table*}[!t]
 \caption{Test scores (mean accuracy $\pm$ standard deviation) of EEGNet and the proposed model.}
 \begin{center}
 \begin{tabular}{ |c|c|c|c| }
   \hline
   Methods & Subject-level scores  & Epoch-level scores   \\ 
   \hline\hline
   EEGNet  & 0.610 $\pm${0.162} & 0.578 $\pm${0.137} \\ 
   \hline
   Proposed Model  & 0.951 $\pm${0.110} & 0.937 $\pm${0.134} \\
   \hline
 \end{tabular}
 \end{center}
 \label{tab:compare_result}
 \end{table*}

\begin{figure*}[!h]
    \centering   
    \begin{subfigure}[b]{0.49\textwidth}
    \includegraphics[width=1\textwidth]{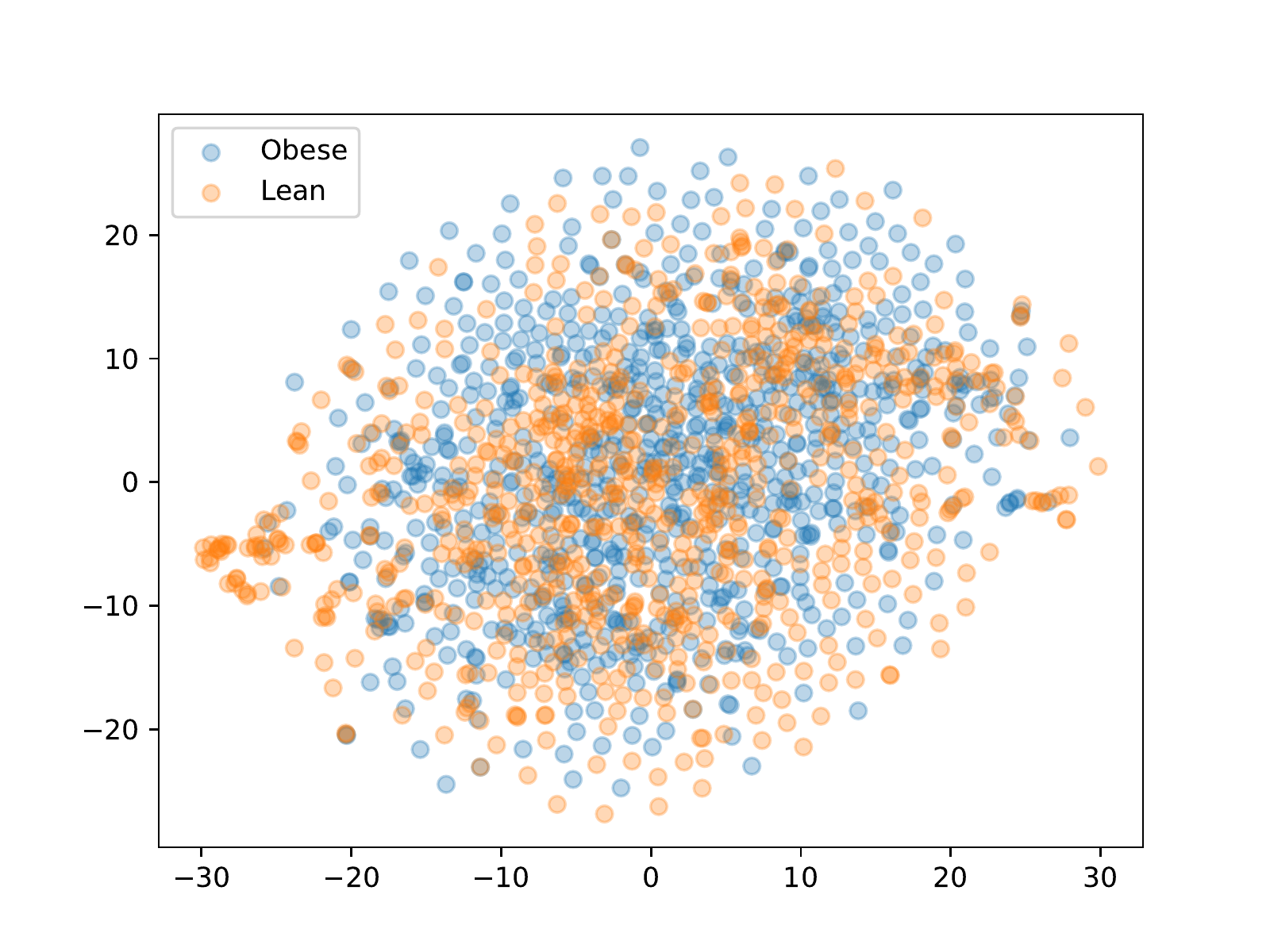}
    \caption{Visualization of features learned by EEGNet.}
    \label{fig:mesh1}
    \end{subfigure}
    \hfill
    \begin{subfigure}[b]{0.49\textwidth}
    \includegraphics[width=1\textwidth]{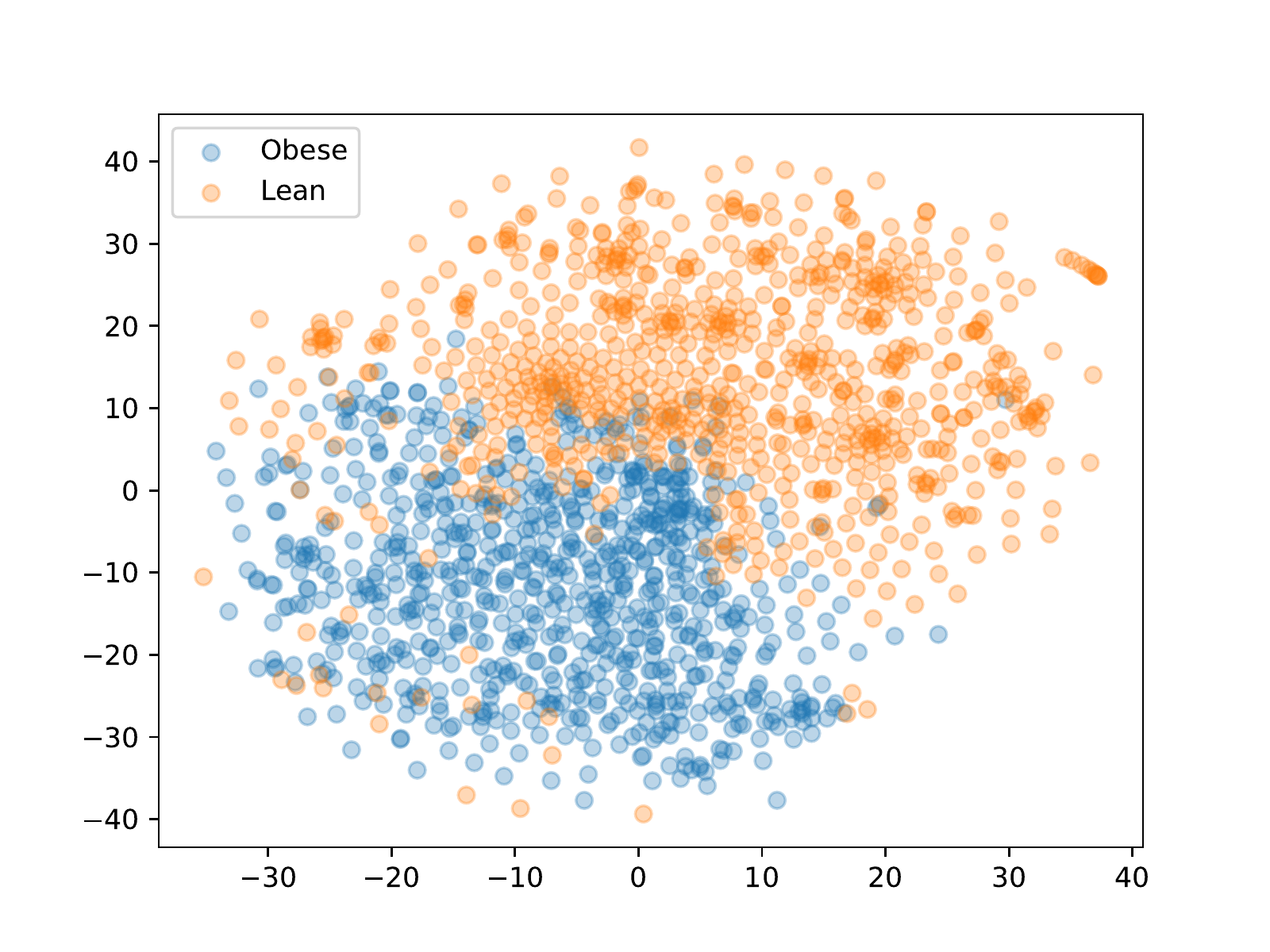}
    \caption{Visualization of features extracted by VAE.}
    \label{fig:mesh2}
    \end{subfigure}      
    \begin{subfigure}[b]{0.49\textwidth}
    \includegraphics[width=1\textwidth]{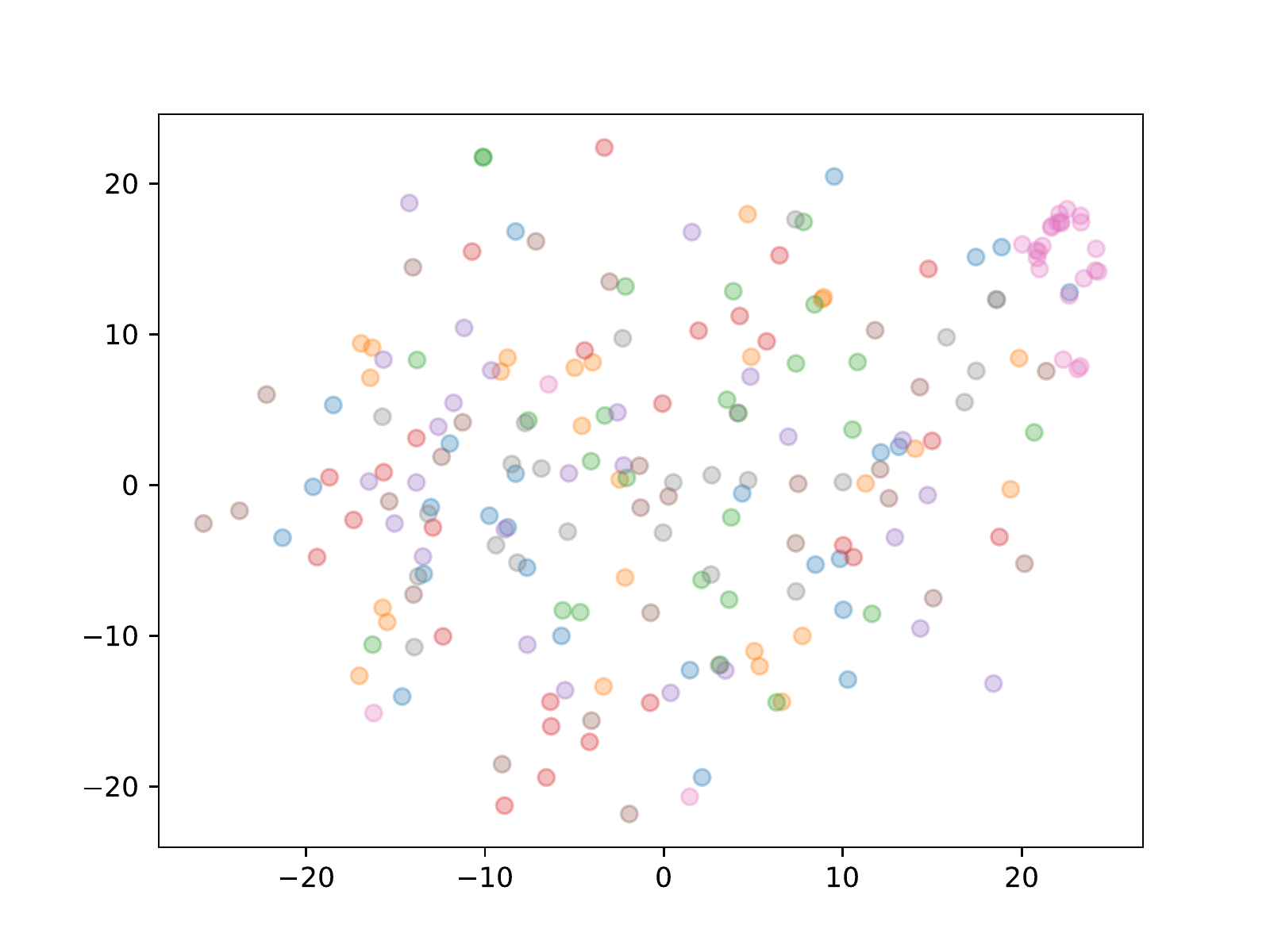}
    \caption{Subject-level view of features learned by EEGNet.}
    \label{fig:mesh3}
    \end{subfigure}  
    \begin{subfigure}[b]{0.49\textwidth}
    \includegraphics[width=1\textwidth]{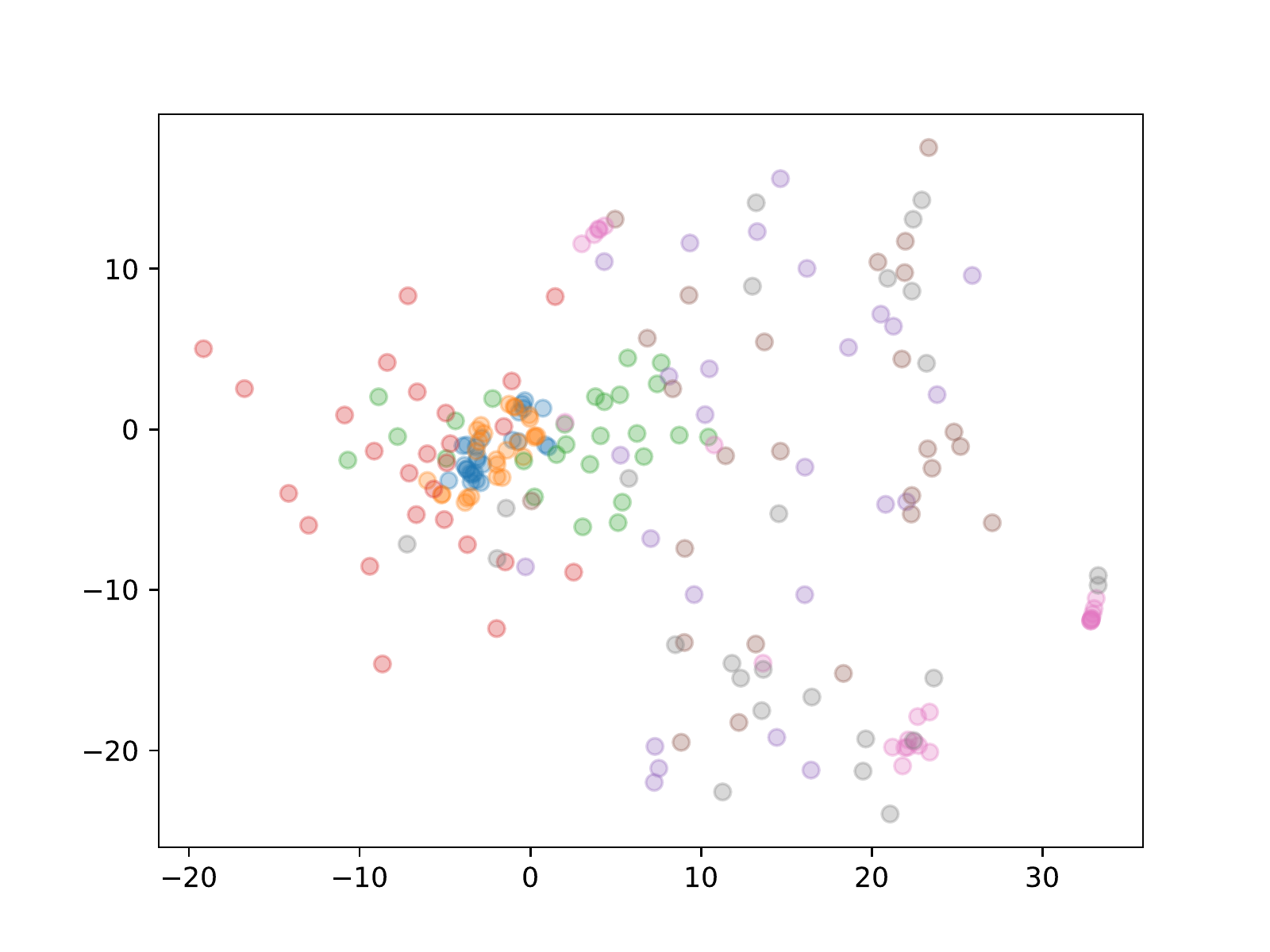}
    \caption{Subject-level view of features learned by VAE.}
    \label{fig:mesh4}
    \end{subfigure}    
\caption{Comparison of 2-D visualizations generated on the learned features using t-SNE. Top row: 2-D projection of features learned by EEGNet (a) and those learned by VAE (b). Obese data samples are represented in blue and lean data samples are represented in orange. Clear separability can be observed from the feature set extracted by the proposed VAE. Bottom row: 2-D projection of features from 6 randomly selected subjects (3 obese subjects and 3 lean subjects) by EEGNet (c), and by VAE (d) . All data points that belong to one subject are in the same color.  Subject-based clusters can be observed in the VAE feature set.}
\label{fig:tsne}
\end{figure*}

As an extensively used model for EEG classification tasks, we chose EEGNet as the baseline model to compare our model with~\cite{lawhern_eegnet_2018}. We directly applied the EEGNet on the raw EEG dataset and the test scores obtained are listed in Table ~\ref{tab:compare_result}. Our proposed model demonstrated its efficiency by significantly improved the classification score from 0.578 to 0.937 and from 0.610 to 0.950 on epoch-level classification and subject-level classification, respectively. 
The Mann-Whitney U test was then conducted and reported a $p$-value of 0.0003 for the subject-level test score and 0.0005 for the epoch-level test score. The $p$-values suggest that there is a significant difference between the two lists of test scores obtained using the EEGNet and our proposed VAE model.


\subsection{Visual comparison of feature representations}

To elaborate on the results, we applied a non-linear dimension reduction technique named t-distributed stochastic neighbor embedding (t-SNE)~\cite{van2008visualizing} to project the high dimensional feature sets learned by the EEGNet and the proposed VAE into 2-D Euclidean spaces. We chose t-SNE here as the dimension reduction method due to it can well preserve the local structure of high dimensional data by minimizing the difference between high dimensional and low dimensional data joint distributions~\cite{birjandtalab_nonlinear_2016}. 

The projected 2-D feature sets that have been learned by the EEGNet and the proposed VAE are shown in Fig. ~\ref{fig:mesh1} and Fig. ~\ref{fig:mesh2}, respectively. It can be seen that features extracted by the proposed VAE are distributed in a well-separated manner  between the obese group and the lean group.


\subsection{Quantitative comparison}
To assess the discriminant ability of feature representations, here we introduce an impurity based measure. Suppose a $D$-dimension, $N$-entry feature set $\mathbf{X}$. Denote the value set of the $i$-th attribute as $X_i$, $i=1,\cdots,D$. Using the idea of decision tree, we seek an optimal threshold $\tau$ that splits the values in $X_i$ into two value subsets with minimum impurity:
\begin{equation}
    \begin{array}{l}
X_i^L = \{x_{ij}|x_{ij}<\tau, j=1,\cdots,N \} \\
X_i^R = \{x_{ij}|x_{ij}\ge\tau, j=1,\cdots,N\}
\end{array}
\end{equation}

As we are dealing with a two-class problem, the impurity can be easily calculated using the Gini index. Suppose within a subset $S$, $p$ is the probability of an instance $x$ belonging to Class 1, the Gini index is 
\begin{equation}
    G(S)= p(1-p).
\end{equation}
Hence we define a ``dichotomy impurity'' for the $i$-th attribute based on the minimal weighted average of impurities of the two subsets generated by the best ``cut'':
\begin{equation}
    DI_i=\min_\tau\left( \frac{|X_i^{L}|}{|X_i|}G(X_i^{L})+\frac{|X_i^R|}{|X_i|}G(X_i^R)\right),
\end{equation}
where $|.|$ indicates cardinality. In other words, $DI_i$ indicates the purest dichotomy we can get on attribute $i$. The overall separability of the feature representation can be roughly indicated by the average $DI$:
\begin{equation}
    DI=\sum_i DI_i/D.
\end{equation}
The smaller $DI$ is, the better separability we can achieve. 

\begin{figure}
\centering
\includegraphics[width=\columnwidth]{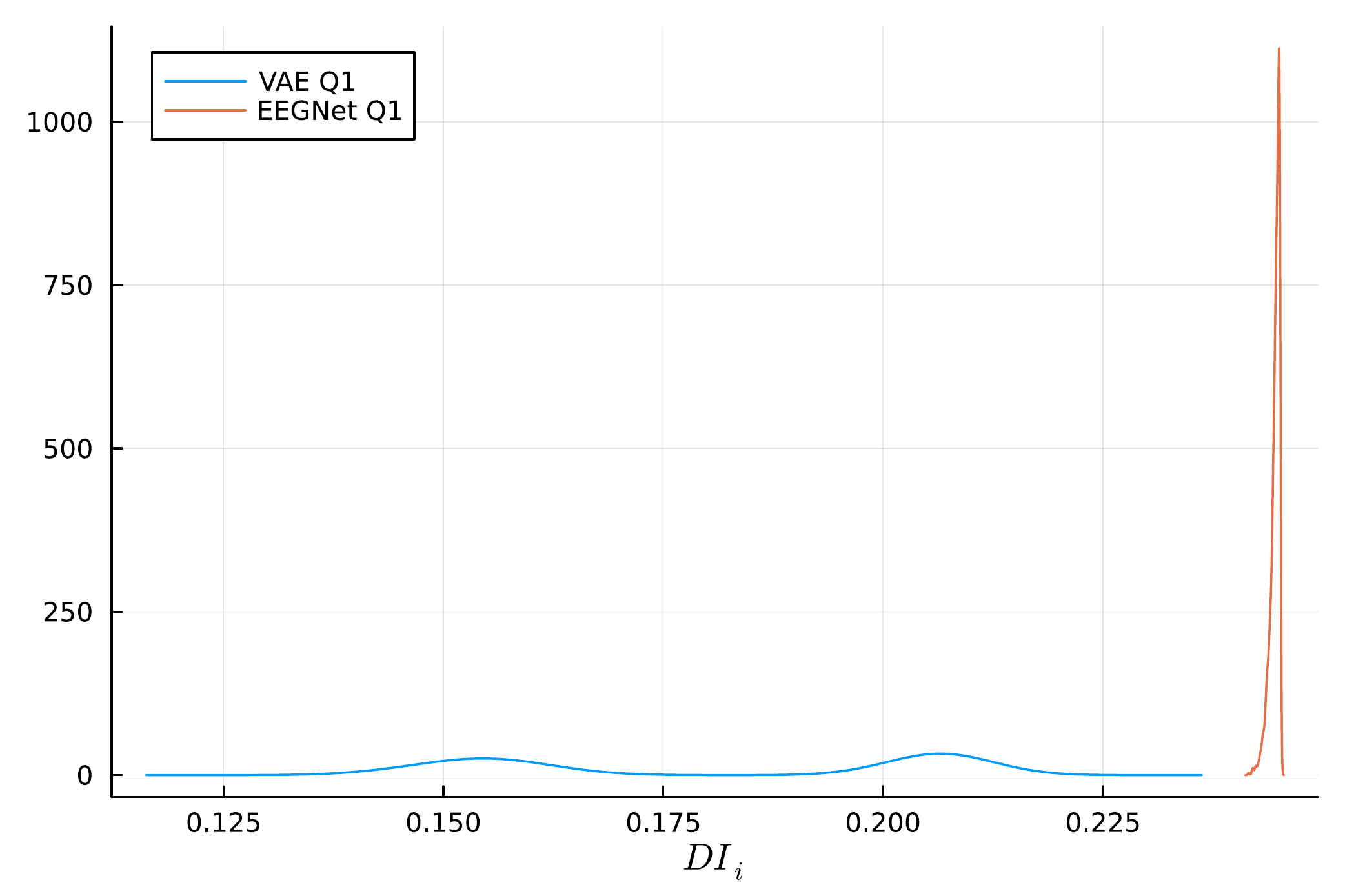}
\caption{Comparison on the first quantile of the dichotomy impurity scores from both feature schemes.}
\label{fig:DI}
\end{figure}

Applying the $DI$ measure to the learned features, we have $DI=0.247$ for the EEGNet features, and $DI=0.220$ for the VAE features. As it is usually the good features that drive the performance of a classifier, we compare the first quantile of these two feature schemes, i.e. the top 25\% of features having the lowest $DI_i$ values, as shown in Fig.~\ref{fig:DI}. Clearly the $DI$ values of the VAE features are well separated from those of EEGNet, hence are much more promising. 

\subsection{Discussion}
One potential reason that explains the good separability in the VAE extracted feature set which leads to the much enhanced performance of our model is that resting state EEG data is highly subject-dependent. By using the proposed VAE to compute a probabilistic latent feature representation, subject-independent elements contained in the raw signals can be extracted whereas subject-dependent elements can be filtered out. 
This idea is illustrated in Fig. ~\ref{fig:mesh3} and Fig. ~\ref{fig:mesh4}. Here we visualized the t-SNE projected data from 3 randomly selected obese subjects and 3 randomly selected lean subjects. All the data points that belong to one subject are represented in the same color. A good cross-subject separability can be observed in the VAE extracted feature set, for instance, the data points that belong to the subjects in orange and blue are well clustered. On the contrary, such a pattern is not shown in the visualized feature set learned by EEGNet. 




\begin{figure}[!t]
     \centering
     \begin{subfigure}[b]{\columnwidth}
         \centering
        \includegraphics[width=\textwidth]{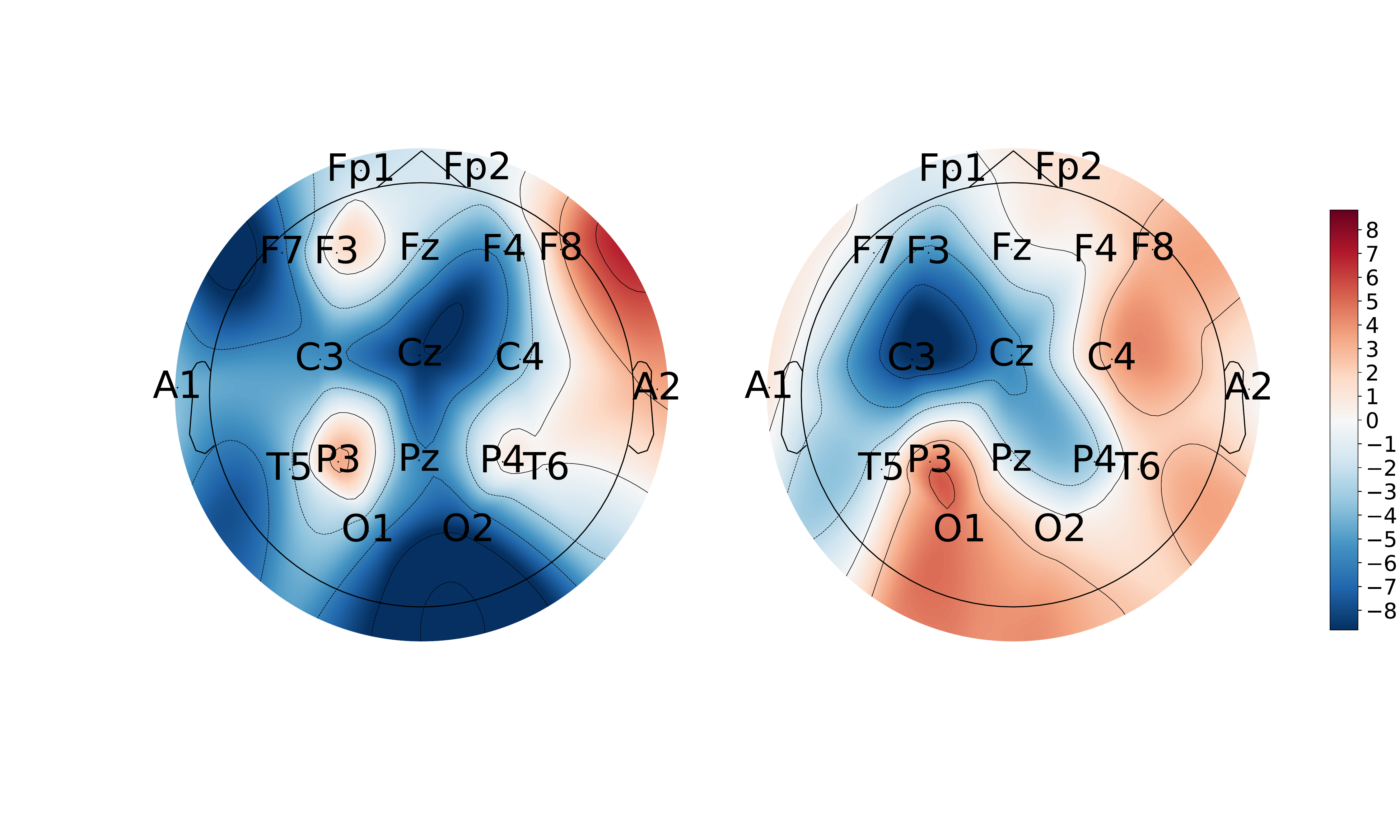}
         \caption{Average spatial patterns for the lean group (left) and the obese group (right). A higher color intensity indicates a higher channel importance.}
         \label{fig:topo1}
     \end{subfigure}
     \begin{subfigure}[b]{\columnwidth}
\includegraphics[width=0.7\textwidth]{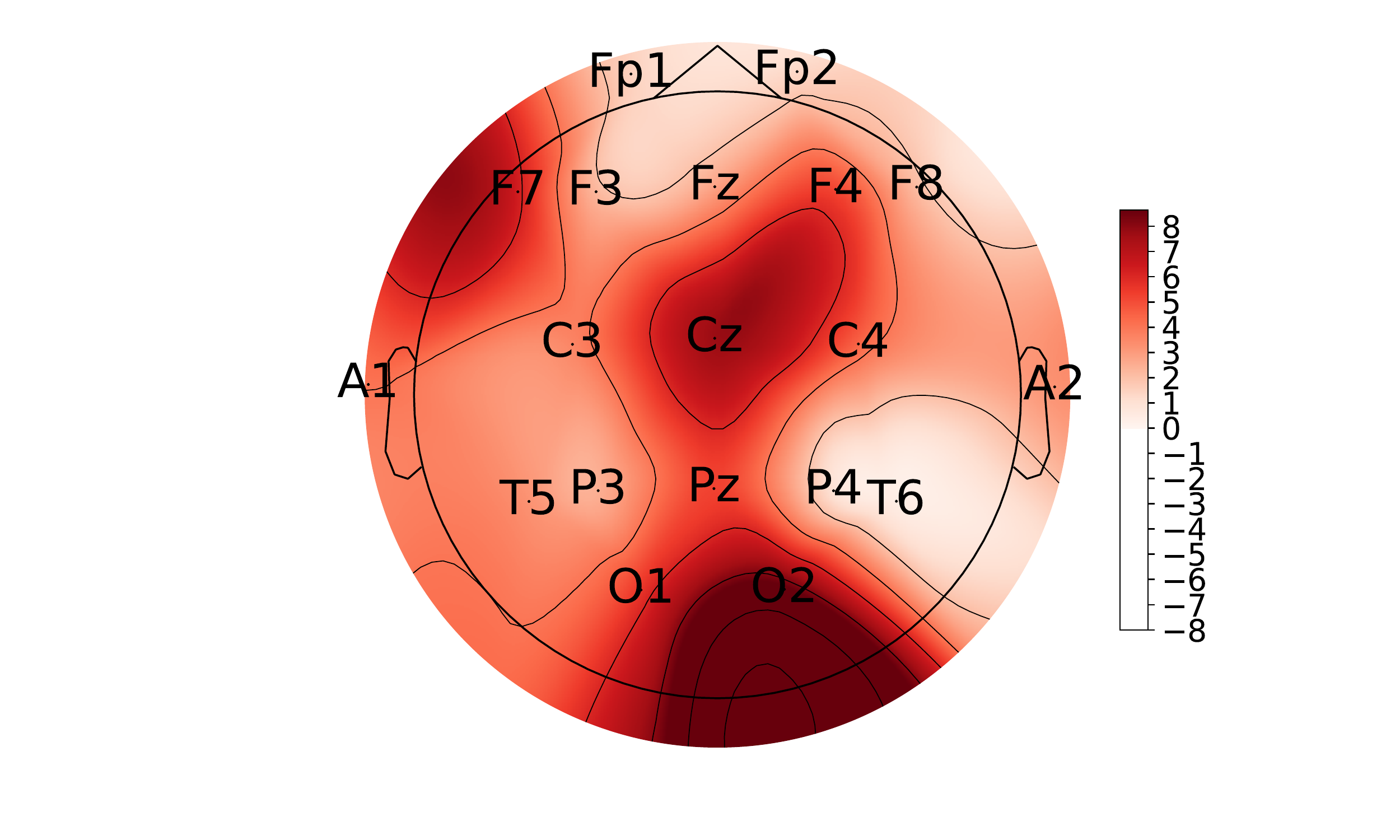}
         \caption{Absolute value of the average difference in each channel between the obese group and the lean group}
         \label{fig:topo2}
     \end{subfigure}
\caption{Visualization of latent features shown as average spatial patterns.}
\end{figure}

Moreover, we infer that EEGNet tends to tweak the learned features toward its gradient results obtained from the classification stage, because it learns in a supervised way. However, as an unsupervised model, VAE learns features with the main goal of reconstructing the input data. Therefore it is less prone to overfit the data in its training process and tends to learn more versatile feature representations. 

On the other hand, insights regarding distinctive spatial patterns between the obese and lean groups can be gained by computing the average value of the second convolution layer output in the encoder of the proposed VAE model. Since in the second convolution layer of the VAE we used filters with a size that equals to $\mathrm{(\#\ of\ channels)}\times 1$,  we can visualize the filter outputs. A larger value in the output feature map potentially suggests a higher importance of the corresponding brain region, furthermore, indicating that the EEG data recorded in that channel contain less noise and therefore makes it easier for the classifier to distinguish between obese and lean brain signals. 


The overall (averaged) spatial patterns learned by the proposed the VAE of the obese group and the lean group are visualized in Fig. ~\ref{fig:topo1}. 
Here, the channel importance is represented by the color intensity, which means both a very dark region and a very reddish region are considered important. An opposite color pattern between two brain regions potentially suggests that the two regions contribute oppositely to the VAE's learning process. 
For a more intuitive demonstration of the channel importance for the classification problem, we computed and visualized the absolute value of the average difference in each channel between the obese group and the lean group in Fig. ~\ref{fig:topo2}, and this allows us to see the net contribution of each channel. Here a darker color simply indicates a higher channel importance.
It can be observed that O2 and O1 appear to be the most significant channels identified by the proposed VAE, followed by F7 and Cz. T6, F8, Fp2, and Fp1 are considered as having only minimal contribution toward the classification.


\section{Conclusion}
\label{sec:conc}
In this study, we have investigated obesity-related brain activities by examining resting state EEG data using a deep learning-based framework. Specifically, we proposed using a VAE to compute the latent representations of raw EEG signals. The features extracted by the VAE are then fed into a 1-D CNN for the final classification. Distinctive spatial patterns learned by the proposed VAE between obese brains and lean brains are also discussed. 
Moreover, we demonstrated the instrumental role of VAE in resting state EEG classification tasks by showing that using the learned latent representations instead of what EEGNet learns from raw EEG signals, it can significantly reduce the subject-dependency and achieves much improved classification performance. The superiority of the VAE features is also reflected by visualizations using t-SNE, and by a dichotomy impurity measure we introduced as a quantitative indicator of class separability. 

For future work, we intend to bring more discussion regarding the neurological aspects of the spatial patterns that have been learned by the proposed model, as well as improve the interpretability of our proposed model by including insights regarding temporal and spectral information.

\bibliographystyle{IEEEtran}
\bibliography{IJCNNbib}

\end{document}